\documentclass[10pt,twocolumn,letterpaper]{article}

\usepackage{cvpr}              

\definecolor{cvprblue}{rgb}{0.21,0.49,0.74}
\usepackage[pagebackref,breaklinks,colorlinks,allcolors=cvprblue]{hyperref}

\def\paperID{62} 
\def\confName{CVPR}
\def\confYear{2025}
\usepackage[accsupp]{axessibility}
\usepackage{multirow}
\usepackage{graphicx}
\usepackage{adjustbox}
\usepackage{float}
\usepackage{pifont}
\usepackage{svg}
\usepackage{algorithm}
\usepackage{algpseudocode}
\usepackage{tikz}
\usepackage{tabularx}
\usepackage{xcolor}

\usepackage{xcolor}
\usepackage{colortbl}
\definecolor{rank_red}{HTML}{FFB3B3}
\definecolor{rank_orange}{HTML}{FFD9B3}
\definecolor{rank_yellow}{HTML}{FFFFB3}
\newcommand{\best}[1]{{\cellcolor{rank_red} #1}}
\newcommand{\sbest}[1]{{\cellcolor{rank_orange} #1}}

\title{SportMamba: Adaptive Non-Linear Multi-Object Tracking\\ with State Space Models for Team Sports}

\newcommand{\proposed}{\texttt{SportMamba}}

\author{Dheeraj Khanna ~~~~~~~~~ Jerrin Bright ~~~~~~~~~ Yuhao Chen ~~~~~~~~~ John S. Zelek\\
University of Waterloo, Waterloo, Ontario, Canada\\
{\tt\small {\{d25khann, jerrin.bright, yuhao.chen1, jzelek\}}@uwaterloo.ca}
}

\begin{document}
\maketitle
\begin{abstract}
Multi-object tracking (MOT) in team sports is particularly challenging due to the fast-paced motion and frequent occlusions resulting in motion blur and identity switches, respectively. Predicting player positions in such scenarios is particularly difficult due to the observed highly non-linear motion patterns. Current methods are heavily reliant on object detection and appearance-based tracking, which struggle to perform in complex team sports scenarios, where appearance cues are ambiguous and motion patterns do not necessarily follow a linear pattern. To address these challenges, we introduce {\proposed}, an adaptive hybrid MOT technique specifically designed for tracking in dynamic team sports. The technical contribution of {\proposed} is twofold. First, we introduce a mamba-attention mechanism that models non-linear motion by implicitly focusing on relevant embedding dependencies. Second, we propose a height-adaptive spatial association metric to reduce ID switches caused by partial occlusions by accounting for scale variations due to depth changes. Additionally, we extend the detection search space with adaptive buffers to improve associations in fast-motion scenarios. Our proposed technique, {\proposed}, demonstrates state-of-the-art performance on various metrics in the SportsMOT dataset, which is characterized by complex motion and severe occlusion. Furthermore, we demonstrate its generalization capability through zero-shot transfer to VIP-HTD, an ice hockey dataset.
\end{abstract}    
\section{Introduction}
\label{sec:intro}
Multi-Object Tracking (MOT) is a key task in computer vision that is traditionally tackled by a series of tasks, e.g., object detection \cite{yolox, detr, center-net, faster-rcnn}, appearance-based Re-ID \cite{fastreid, zhou2021osnet}, motion prediction \cite{sort, MOTR, bytetrack, deepsort}, and temporal associations \cite{deep-eiou, cbiou}. Research on MOT has been conducted on various practical use cases, including sports \cite{sportsmot, soccernet}, dancing scenes \cite{dancetrack}, and driving scenes \cite{bdd100k, kitti}.

\begin{figure}[t]
    \centering
    \includegraphics[width = 0.49\textwidth]{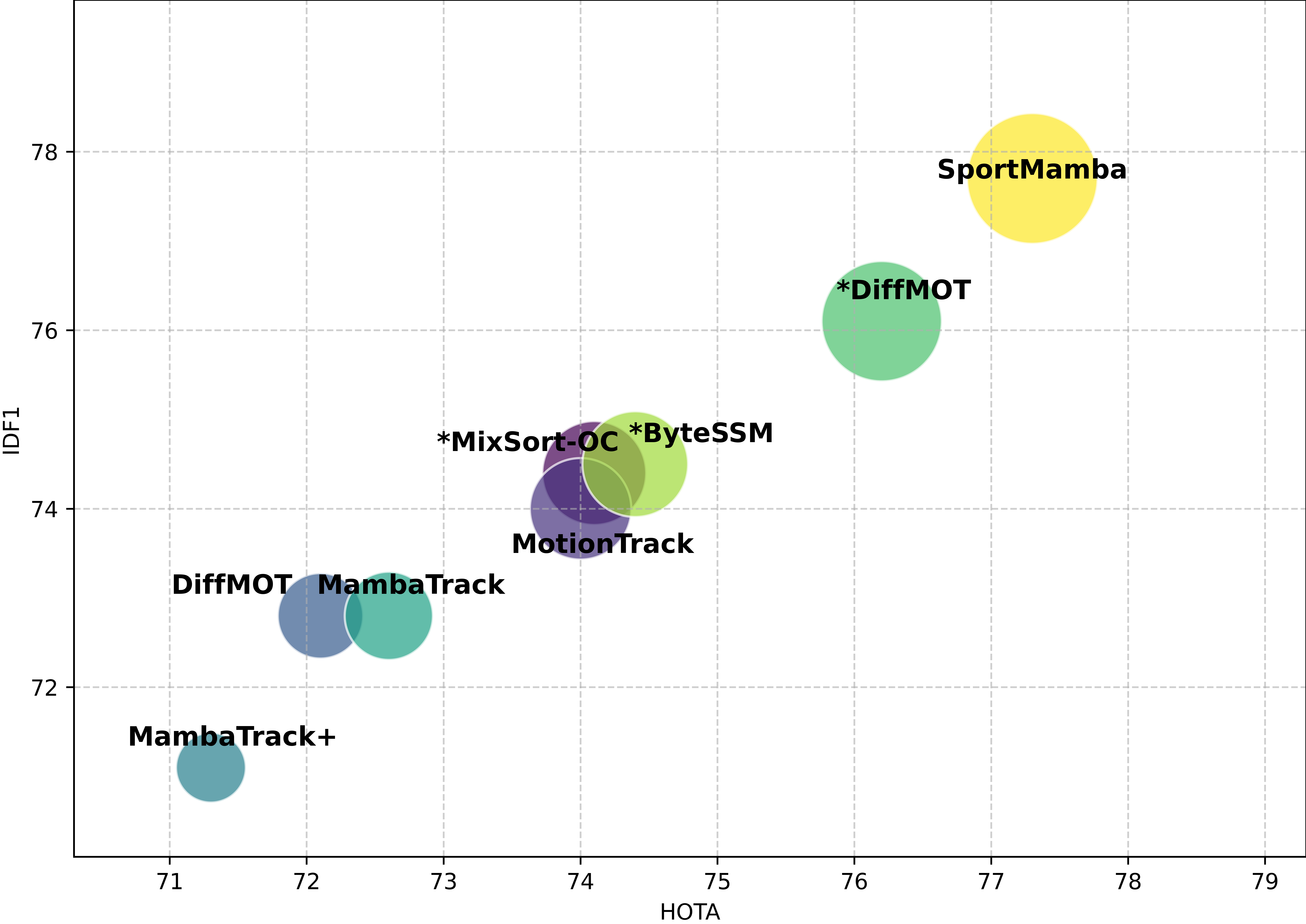} %
    \vspace{-20px}
    \caption{\textbf{Comparison of HOTA and IDF1 score on SportsMOT dataset with recent learning-based methods.} State-of-the-art (SOTA) performance compared to prior works is demonstrated with the proposed {\proposed} model. Methods marked with * use training and validation data for the object detector.}
    \label{fig:comparison-hoto}
    \vspace{-10px}
\end{figure}

The primary objective of MOT is to temporally track objects while maintaining unique identities. However, factors such as occlusion, motion blur, and unpredictable motion patterns significantly influence the performance of prior works \cite{dancetrack, soccernet, sportsmot, vip-htd}. These issues are particularly exacerbated in team sports like ice hockey \cite{vip-htd, VATS2023119250}, soccer \cite{sportsmot, soccernet}, and baseball \cite{jerrin_baseball}, where fast-paced movements and frequent switching between players make it even more challenging.

Conventional methods \cite{sort, deepsort, bytetrack} rely predominantly on heuristic-based approaches like the Kalman filter \cite{kalmanfilter} or its variants \cite{particle_filters} and usually struggle with arbitrary non-linear motion \cite{ocsort}, such as highly random skating actions of the ice hockey players. Recently, transformer-based learning approaches have been leveraged for this task \cite{trackformer, MOTR, motiontrack} but are not suitable for real-time tracking in sports due to their computational complexity \cite{motr-v3, memotr}.

Recently, State-Space Models (SSMs) such as Mamba \cite{mamba} have emerged as an alternative that demonstrates strong sequence modeling capabilities in linear time. Preliminary attempts on Mamba for object prediction \cite{mambatrack, mambamot, xie2024robust, trackssm} have demonstrated efficient temporal modeling of player tracklets with comparatively less computational overhead. Although beneficial, literature highlights a key limitation: SSMs compress historical information, which will tend to be unreliable during prolonged occlusion or when nonlinear motion is observed \cite{wen2024rnns}. To address these issues, existing works \cite{xiao2024mambatrack} proposed autoregressive prediction steps to infer the missing states but struggled due to the accumulation of error over time. This will ultimately lead to the fragmentation of trajectories and increased identity switches.

Inspired by these works, we propose {\proposed}, a hybrid learning-based online tracking and association model tailored for tracking fast-moving objects in team sports. The {\proposed} architecture is a four-stage process: First, an off-the-shelf detector is fine-tuned for sports-specific player detection. Next, a mamba-attention motion predictor estimates player positions in subsequent frames. The predicted bounding boxes are then matched with detections from the object detector using a hybrid matching metric, which combines appearance-based Re-ID features with a novel height adaptive IoU metric with extended buffers. Extensive experimentation revealed SOTA performance across different metrics when compared against benchmarked SportsMOT and VIP-HTD datasets (see Figure \ref{fig:comparison-hoto} for a visual comparison). Our contributions can be summarized as follows:

\begin{itemize}
    \item We propose {\proposed}, a hybrid learning-based model for online tracking and temporal association, designed to handle fast-paced, non-linear motion scenarios such as team sports, while efficiently running $\approx$ 30 FPS during inference.
    \item We introduce a novel motion predictor model that integrates Mamba’s state-space modeling with self-attention.
    \item We introduce a height-adaptive IoU with extended buffers for spatial association, leading to robust detection to tracklet matching.
    \item Extensive experimentation of {\proposed} on the SportsMOT dataset demonstrates SOTA performance across multiple tracking metrics. 
    \item {\proposed} exhibits superior zero-shot generalizability on the VIP-HTD ice hockey dataset, outperforming prior works across most metrics. 
\end{itemize}

\section{Related Works}
\label{sec:lit}

\subsection{Multi Object Tracking}
 MOT is broadly classified into Tracking-by-Detection (TBD) and Joint-Detection and Tracking (JDT) paradigms, where both have a commonality of object detection as their first critical step. MOT has seen significant progress in recent years with the advancement of SOTA object detection models \cite{yolox, yolox, center-net, detr}. The next 2 steps in MOT are motion prediction and data association respectively.
\subsubsection{Motion Models in MOT}
 Motion Prediction includes estimating the state of the object in the next frame given the information about its current state using a motion model. SORT \cite{sort} was the first method to use Kalman Filter (KF) \cite{kalmanfilter} as a motion model for the prediction task. This work was followed by several other works \cite{deepsort, bytetrack, ocsort, deep-oc-sort, botsort, strongsort}. Methods such as GIAOTracker \cite{giaotracker} and OC-SORT \cite{ocsort} further modify KF to incorporate non-linear motion where traditional KF-based methods lack due to its linear velocity assumption. Methods such as BotSORT \cite{botsort} and Deep OC-SORT \cite{deep-oc-sort} utilize camera motion compensation and update predictions. Despite these advancements, KF-based motion models rely on predefined hyperparameters and cannot be truly generalized for different applications of MOT. 

 Learning-based motion models are required to mitigate the limitations of KF by being data-driven that can learn an object's state by learning its past movements. These models capture complex temporal dependencies and non-linear motion patterns of an object for the task of prediction. ArTIST \cite{artist} proposes to solve the object prediction task by creating a nonlinear stochastic motion model that uses probability distribution to estimate the likelihood of the object in the next frame. MotionTrack \cite{motiontrack} utilizes a Transformer \cite{attentionisallyouneed} architecture and a Dynamic MLP \cite{motiontrack} to predict the next state of the object. DiffMOT \cite{diffmot} and DiffusionTrack \cite{diffusiontrack} treat motion prediction as a generative task. MambaTrack \cite{mambatrack} proposes a vanilla-mamba \cite{mamba} architecture for motion prediction while also incorporating the hidden states for data association. TrackSSM \cite{trackssm} employs an encoder-decoder architecture, where the encoder utilizes Mamba-based layers to capture motion representations while the decoder integrates a Flow-SSM module to enable the temporal autoregression of bounding boxes. These methods improve prediction accuracy through data driven modeling, however, their effectiveness is limited wihout robust data association.
 \vspace{-5px}

 \begin{figure*}[t]
    \centering
    \includegraphics[width = \textwidth]{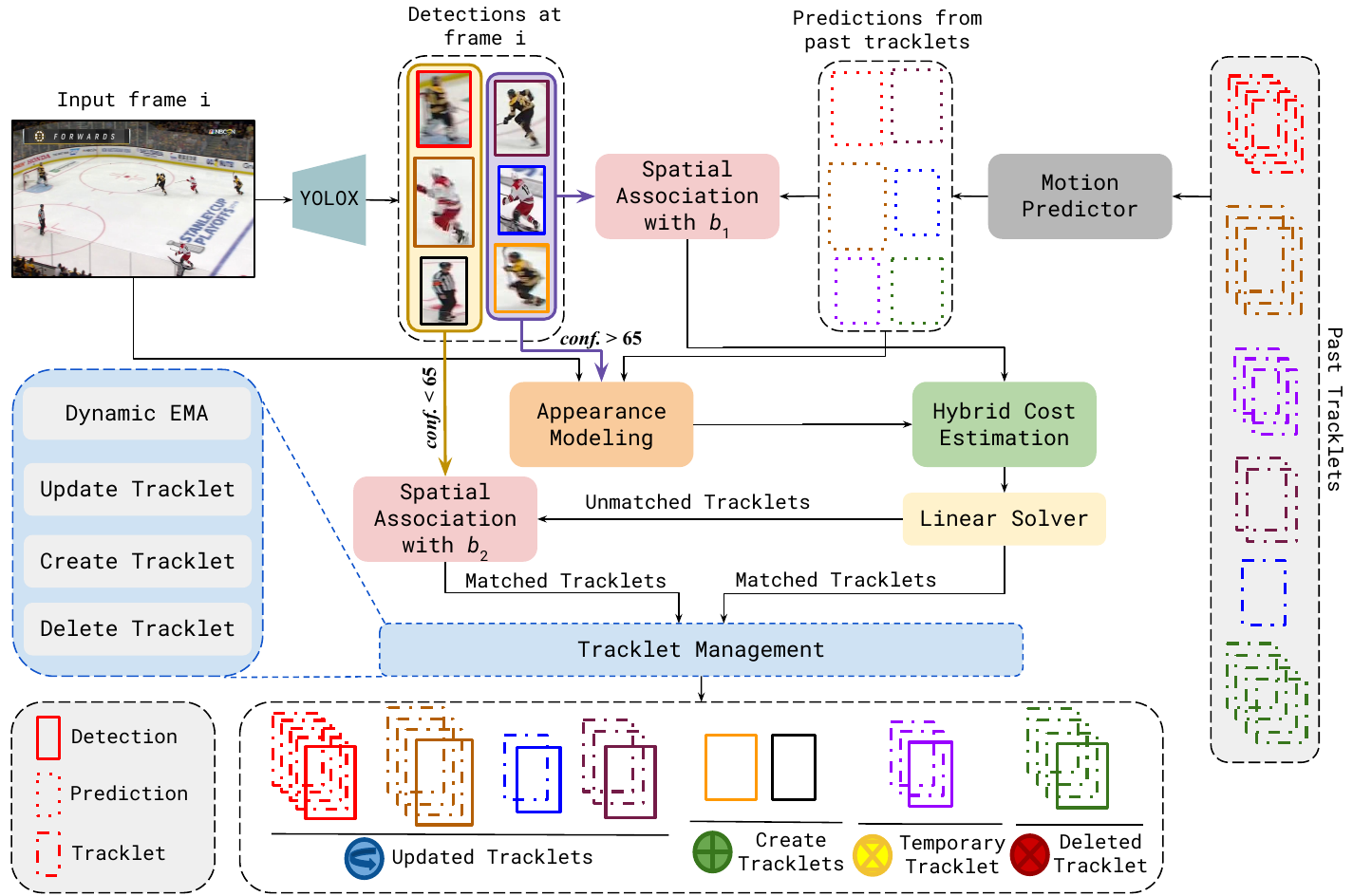} %
    \vspace{-20px}
    \caption{\textbf{Overview of {\proposed}.} The architecture follows a four-stage process: (1) A fine-tuned object detector detects players; (2) A motion predictor estimates future player positions based on past tracklets; (3) A high-confidence association using a hybrid matching metric integrating appearance modeling and a hybrid cost estimated by a height-adaptive IoU with extended buffers. (4) Low-confidence and unmatched matches from (3) are reassociated using a relaxed IoU without appearance matching. {\proposed} concludes with tracklet management- update, create, delete operations and dynamic EMA of the updated tracklets.}
    \label{fig:sportmamba}
    \vspace{-10px}
\end{figure*}

\subsection{Multi-Object Tracking for Sports}
Team sports introduce challenges such as similar appearances due to having the same jersey, fast-paced games, and frequent camera motions \cite{sportsmot, soccernet, vip-htd, hockeymot}. Recent methods \cite{sportsmot, deep-eiou, cbiou} primarily focus on improving the data association stage using improved spatial and appearance-based matching. HockeyMOT \cite{hockeymot} utilizes a Message Passing network in a graph-based structure to enhance association on ice hockey data. MixSort \cite{sportsmot} utilized an attention-based mechanism "MixFormer" \cite{mixformer} for getting the appearance-features and combine them with other motion models such as ByteTrack \cite{bytetrack} and OC-SORT \cite{ocsort}. CBIoU \cite{cbiou} and Deep EIoU \cite{deep-eiou} extended the bounding box sizes to compensate for fast-motion scenarios. Deep HM-SORT \cite{deephmsort} modifies the cost matrix using a harmonic mean between the spatial and appearance cost matrix. Despite these advancements, combining efficient data association with robust motion prediction requires further exploration to address challenges like fast-moving players, overlaps, and camera shifts.


\section{Methodology}
\label{sec:method}
This section begins with the Problem Formulation in Sec. \ref{subsec:problem_formulation}, followed by our proposed attention-Mamba-based motion model in Sec. \ref{subsec:motion-prediction-model}. Sec. \ref{subsec:tracklet-association} details our hybrid data association approach, including height adaptation and extended IoU, while Sec. \ref{subsec:track-management} outlines the overall track management process within the tracking pipeline.

\begin{figure*}[t]
    \centering
    \includegraphics[width=0.9\linewidth]{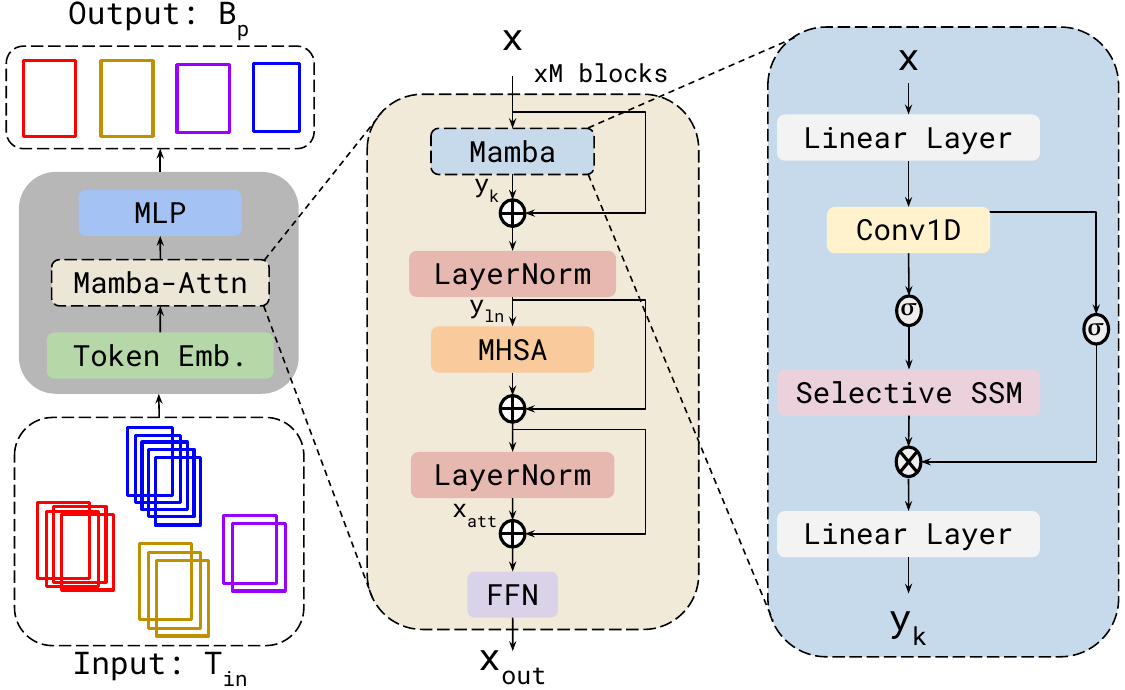}
    \vspace{-10px}
    \caption{\textbf{Overview of the Motion Prediction Model.} The model first encodes past object trajectories ($T_{in}$) using a Token Embedding Layer, followed by a Mamba-Attention Encoder that models long-range motion dependencies using a discretized state-space representation. An MHSA Block refines the Mamba-encoded features followed by FFN for feature transformation. After passing through $M$ stacked Mamba-Attention blocks, the final representation is fed to a prediction head (MLP) to output the bounding box for the next frame.}
    \label{fig:architecture}
    \vspace{-3mm}
\end{figure*}

\subsection{Problem Formulation}
\label{subsec:problem_formulation}
{\proposed} follows an online MOT approach and adheres to the TBD paradigm. Given an input image, we employ a detector to output $(B_d, s_t)$ where $B_d$ are the detected bounding boxes for all objects in a frame, defined as $B_d = \{x_t, y_t, w_t, h_t\}$. Here, $(x_t, y_t)$ denotes the top-left corner coordinates, while $w_i$ and $h_i$ represent the width and height of the bounding box. The confidence score for each detection is given by $s_t \in [0, 1]$. 

For motion prediction, an input tracklet of $l$ is defined as  $T_{in}= \{B^{t - l}_d, B^{t - l + 1}_d, ... B^{t-1}_d\}$, where each bounding box $B_d^i$ is represented as $\{x_d^i, y_d^i, w_d^i, h_d^i\}$. The motion predictor model takes in $T_{in}$ as input length $l \in (2, w)$ and predicts the state of the bounding box $B_p$ at time $t$. Here, $w$ is the maximum tracklet length used for training.

\subsection{Motion Prediction Model}
\label{subsec:motion-prediction-model}
\textbf{Token Embedding.} The input trajectory consisting of past bounding box sequence is transformed into a high-dimensional embedding using a linear projection layer.

\noindent \textbf{Mamba-Attention Encoder.} Mamba is a SSM technique which, upon discretization with Zero-Order Hold (ZOH) assumption over timestep $\Delta$, can be formulated as:

\begin{equation}
\begin{aligned}
    h'_k &= A h_{k-1} + B x_k \\
    y_k &= C h_k    
\end{aligned}
\end{equation}

\noindent where, $A$, $B$ and $C$ are dynamically adjusted parameters based on input features $x_k$ and hidden state $h_k$ at time $t$. This enables recursive updating of the hidden states to model motion dependencies effectively.

To model non-linear motion effectively by implicitly focusing on relevant dependencies, we add a Multi-Head Self-Attention (MHSA) block to the mamba features $y_k$. The MHSA mechanism can be written as shown in Equation \eqref{eq:MHSA}.
\vspace{-5px}
\begin{equation}
\begin{aligned}
    y_{ln} &= LayerNorm(y_k) \\
    x_{att} &= LayerNorm(MHSA(y_{ln}) + y_{ln})
\label{eq:MHSA}
\end{aligned}
\end{equation}

\noindent where MHSA has $L$ heads, where each head is denoted as:
\vspace{-5px}

\begin{equation}
    head_i = {Softmax} (\dfrac{Q^i (K^i)^T}{\sqrt{d}})V^i \label{eq:zs}
\end{equation}

Here, $y_{ln}$ is linearly transformed to obtain queries $Q$, key $K$ and value $V$. The attended features ($x_{att}$) are then fed to a feed-forward network (FFN), which consists of GeLU activation and linear layers, represented as:
\vspace{-5px}
\begin{equation}
    x_{ff} = Linear(GeLU(Linear(x_{att})) + x_{att})
\end{equation}

\noindent \textbf{Prediction Head.} The final representation $x_{out}$ is obtained from the last mamba-attention block, i.e., $x_{out} = x^M_{ff}$, and is then fed to a prediction head, which consists of a linear layer followed by a sigmoid activation function. This can be formulated as as shown in Equation \eqref{eq:pred_head}
\vspace{-5px}

\begin{equation}
    B_p = \sigma(Linear(x_{out})) \label{eq:pred_head}
\end{equation}


\subsection{Tracking and Association}
\label{subsec:tracklet-association}
The association consists of two stages: a high-confidence association (HCA) and a low-confidence association (LCA). The primary objective of HCA and LCA is to establish correspondences between the detections from the object detector and the motion predictions from the mamba-attention block. HCA first associates detections $B_d$ with predicted motion labels $B_p$ based on high-confidence \textit{spatial} and \textit{appearance-based cues}, ensuring stable identity tracking. However, this might prematurely drop detections due to the high-confidence threshold. Thus, LCA provides an additional matching step for low-confidence detections, leveraging only a stricter spatial association constraints to recover tracklets that temporarily lost detections.  

\subsubsection{High-Confidence Association} 

Inspired by the matching strategy of ByteTrack \cite{bytetrack}, we introduce a hybrid association strategy that combines spatial association with appearance-based feature extraction. 

\noindent \textbf{Spatial association.} The conventional IoU metric predicts the match between the predicted bounding box $B_p$ and detected bounding box $B_d$, but is overly strict, leading to over-penalization by missing associations with slight shifts. Thus, following previous work \cite{cbiou, deep-eiou}, we extend the bounding box with an expansion factor or buffer $b_1 \in [0, 1]$. This technique is termed as Extended IoU (EIoU) and is computed in Equation \eqref{eq:eiou_eq}.
\vspace{-5px}
\begin{equation}
    EIoU = \frac{|B'_p \cap B'_d|}{|B'_p \cup B'_d|}
\label{eq:eiou_eq}
\end{equation}

\noindent where, $B'_d = \{x_t - 0.5bw_t, y_t - 0.5bh_t, w_t + bw_t, h_t + bh_t\}$ and $B'_p = \{\hat{x}_t - 0.5b\hat{w}_t, \hat{y}_t - 0.5b\hat{h}_t, \hat{w}_t + b\hat{w}_t, \hat{h}_t + b\hat{h}_t \}$.

However, relative height differences are often ignored in these matching metrics. We argue that, especially in sports scenarios, height-aware matching is necessary to tackle the issue with varying depths with the scale. Thus, we introduce Height-Adaptive EIoU (HA-EIoU), defined in Equation \eqref{eq:heiou}.
\vspace{-5px}
\begin{equation}
    \text{HA-EIoU} = \text{HIoU}.\text{EIoU} 
    \label{eq:heiou}
\end{equation}

\begin{figure}[t]
    \centering
    \includegraphics[width=\linewidth]{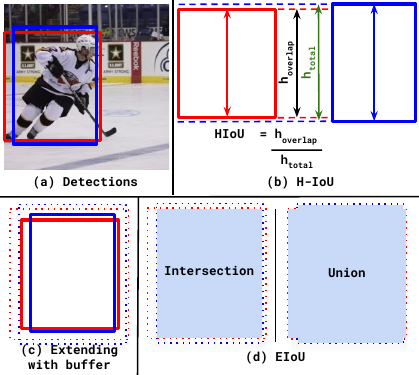}
    \vspace{-20px}
    \caption{\textbf{Visual Representation of HIoU against EIoU.} (a) Visualizes the detection with groundtruth (in \textcolor{blue}{\textbf{blue}}) and prediction (in \textcolor{red}{\textbf{red}}); (b) Representation of the HIoU metric; which is the height overlap by the total height; (c) Extended bounding box represented with dotted lines based on a buffer $b$; and (d) Representation of EIoU using the extended bounding boxes.}
    \label{fig:hiou}
    \vspace{-10px}
\end{figure}

\noindent where, HIoU measures how well the bounding boxes ($B_d$ and $B_p$) align with respect to height, represented in Equation \eqref{eq:hiou}. Figure \ref{fig:hiou} shows a visual representation of HIoU compared to EIoU.
\vspace{-5px}

\begin{equation}
    \text{HIoU} = |  \frac{\min(y^2_t,\hat{y}^2_t) - \max(y^1_t,\hat{y}^1_t)}{\max(y^2_t, \hat{y}^2_t) - \min(y^1_t, \hat{y}^1_t)} |
    \label{eq:hiou}
\end{equation}

\noindent \textbf{Appearance-based feature extraction.} While EIoU and HIoU measure spatial similarity, they struggle when there is considerable occlusion. Thus, we capture appearance features \cite{fastreid} as an additional metric to resolve this ambiguity by differentiating players with color, texture, and shape characteristics even when their spatial positions are similar. 

\noindent \textbf{Hybrid Cost Estimation.} To establish correspondence using appearance and spatial similarity, we compute the cost matrix ($\mathcal{J}_f$). First, the appearance embedding for the tracklet and the detection is taken. Then, the cosine similarity between these embeddings is taken to find the appearance cost ($\mathcal{J}_{reid}$), as shown in Equation \eqref{eq:s_reid}. 
\vspace{-5px}

\begin{equation}
    S_{reid}(i, j) = \frac{e_i^T \cdot e_j^D}{\|e_i^T \| \|e_j^D \|}
    \label{eq:s_reid}
\end{equation}

\noindent where, $e_i^T$ and $e_j^D$ are the feature embeddings of the tracklet and detection, respectively. This results in a $N \times M$ similarity matrix ($S_{reid}$). Then, spatial similarity ($\mathcal{J}_{ssim}$) is computed using Equation \eqref{eq:heiou}. These cues are then integrated using a weighted sum represented as:
\vspace{-5px}
\begin{equation}
\begin{aligned}
    \mathcal{J}_f &= \lambda_{reid} \mathcal{J}_{reid} + \lambda_{ssim} \mathcal{J}_{ssim} \\
    \mathcal{J}_f &= \lambda_{reid} (1 - S_{reid}) + \lambda_{ssim} (1 - \text{HA-EIoU})
\end{aligned}
\end{equation}

Once the cost matrix ($\mathcal{J}_f$) is constructed, a linear assignment solver \cite{Kuhn1955TheProblem} is used to estimate the one-to-one matching between the tracklets and detections. 

\subsubsection{Low-Confidence Association} 
The unmatched tracklets from HCA are fed to the LCA block along with the low-confidence detections. In this step, the spatial association detailed in HCA is performed with stricter constraints- a buffer $b_2$, where $b_2 < b_1$. The appearance-based ReID is, however, not used since detections with lower confidence have a higher tendency for noisy detections or artifacts present in it. Therefore, the matching of the unmatched tracklets and low-confidence detections is done only with HA-EIoU. 

\subsection{Track Management}
\label{subsec:track-management}
The track management step identifies the tracklet's status, handling three key operations: updating existing tracklets, creating new tracklets, and deleting lost tracklets. Additionally, it updates the tracklet features with the new detections. 

\noindent \textbf{Tracklet update.} Once the linear solver assigns a detection to a tracklet, the detection's bounding box is concatenated with the tracklet's existing detections. This will then be used in predicting detections in future frames. 

\noindent \textbf{Tracklet creation.} If a detection does not match any existing tracklet, a new tracklet is initialized. This is particularly frequent in sports, where players move out of the camera's field of view, undergo player substitutions, or if a previously lost tracklet cannot be recovered.

\noindent \textbf{Tracklet deletion.} If a tracklet doesn't have a corresponding detection at the current frame, it is marked as lost. Lost tracklets are temporarily stored in memory, allowing for re-identification if the player reappears. However, if a tracklet remains unmatched within the timeframe, it is permanently deleted from the memory. 

\noindent \textbf{Dynamic feature updating.} To avoid computational overhead by extracting appearance embedding of the tracklets for each iteration, prior works \cite{strongsort, botsort} utilized Exponential Moving Average (EMA) for updating tracklet features. The updated embedding $e^{t-1}_i$ after matching with a detection is computed as shown in Equation \eqref{eq:updated_feat}.

\vspace{-5px}

\begin{equation}
    e^{t-1}_i = \alpha e^{t-2}_i + (1-\alpha)f^{t-1}_i
\label{eq:updated_feat}
\end{equation}

\noindent where $\alpha$ is the smoothing factor that determines the contribution of the historical embedding $e^{t-1}_i$ relative to the new detection embedding $f^{t-1}_i$. However, in challenging visual scenarios like in ice hockey, where occlusion and motion blur are prevalent, the newly extracted features might not properly reflect a player's true appearance due to the possibility to artifacts. Thus, we adopt a dynamic EMA mechanism, where $\alpha$ is updated based on the detection confidence score $s_t \in [0,1]$. The updated smoothing function ($\alpha_{d}$) is shown in Equation \eqref{eq:dynamic-ema}.
\vspace{-5px}

\begin{equation}
    \alpha_d = \alpha + (1-\alpha)\frac{1-(s_t - \sigma)}{1-\sigma}
    \label{eq:dynamic-ema}
\end{equation}

\noindent where, $\sigma$ is the minimum confidence threshold. The updated embedding formulation with the dynamic smoothing function can be written as:
\vspace{-5px}

\begin{equation}
    e^{t-1}_i = \alpha e^{t-2}_i + (1-\alpha_d)f^{t-1}_i
\end{equation}

\subsection{Objective Function}

The motion predictor is trained using a combination of a variant of L1 loss and IoU loss function. Smooth L1 loss \cite{fast-rcnn} is used to measure the difference between the predicted and groundtruth bounding boxes. This loss function is represented as shown in Equation \eqref{eq:smooth_loss}.
\vspace{-5px}
\begin{equation}
    \mathcal{L}_{L1}^s =
    \begin{cases} 
        \frac{1}{2} (P_t - G_t)^2, & \text{if } |P_t - G_t| < 1, \\
        |P_t - G_t| - \frac{1}{2}, & \text{otherwise}.
    \end{cases}
    \label{eq:smooth_loss}
\end{equation}

Additionally, we utilize Complete IoU (CIoU) denoted as $\mathcal{L}_{ciou}$, which evaluates the center alignment, overlap, and aspect ratio consistency of predicted bounding boxes with the groundtruth. The overall loss function of the motion prediction model is denoted as shown in Equation \ref{eq:overall_loss}.
\vspace{-5px}

\begin{equation}
    \mathcal{L} = \lambda_{l1}^s \mathcal{L}_{L1}^s + \lambda_{ciou} \mathcal{L}_{ciou}
    \label{eq:overall_loss}
\end{equation}
\section{Experimentation}
\label{sec:exp}

\subsection{Implementation Details} 

\textbf{Training Details.} All experimentations were conducted on a single NVIDIA 4090 GPU with 24GB of vRAM. {\proposed}'s motion prediction model was trained for 60 epochs with a batch size of 64 with $M=4$ blocks. The model was trained using the AdamW optimizer with $\beta_1=0.9, \beta_2=0.98$, a learning rate of $10^{-4}$, and a weight decay of $10^{-3}$. The maximum sequence length was set to $w$ during training. $\lambda_{L1}^s$ and $\lambda_{ciou}$ were set to 50 and 1 respectively. Following prior works \cite{bytetrack, ocsort, diffmot, deep-oc-sort, strongsort}, we train a YOLOX \cite{yolox} model for player detection. 


\noindent \textbf{Data augmentation.} Two data augmentation strategies are incorporated: \textit{1) Temporal augmentation}, where a random frame sequence of length between 2 and $w$ is selected from a given tracklet, with padding applied to ensure equal tracklet length of $w$ for batch processing; \textit{2) Spatial augmentation}, which includes random scaling and translation and gaussian noise injection to represent for detection inaccuracies. To prevent excessive distortion, no single tracklet undergoes all transformations simultaneously.

\subsection{Datasets}

\textbf{SportsMOT.} SportsMOT dataset \cite{sportsmot} includes 240 video sequences of three sports (i.e., basketball, soccer, and volleyball) captured in 720p at 25 FPS. It encompasses fast-paced motion with variable speeds and complex, non-linear motion. This dataset serves as a good benchmark to evaluate {\proposed}'s ability to track players across sports.  

\noindent \textbf{VIP-HTD.} VIP-HTD dataset \cite{vip-htd} includes ice hockey broadcast feed for 8 games with 22 sequences in total captured in 720p at 30 FPS. Similar to SportsMOT, VIP-HTD features rapid movements with varying speeds and unpredictable trajectories. However, due to the \textit{faster} pace of the ice hockey game, it exhibits comparably more motion blur. This dataset was chosen to assess the robustness of {\proposed} in generalizing to more extreme conditions.

\subsection{Evaluation Metrics}

To analyze different components of {\proposed}, Higher-order Tracking Accuracy (HOTA) \cite{hota}, Identification F1 score (IDF1) \cite{idf1}, Association Accuracy (AssA), Detection Accuracy (DetA), and Multiple Object Tracking Accuracy (MOTA) \cite{clearmot} are used. Each metric highlights a specific aspect of the performance of {\proposed}- DetA evaluates the detection accuracy irrespective of identity tracking, AssA and IDF1 measure identities association quality, and to evaluate both detection and tracking performance, HOTA and MOTA are used. 

\subsection{Benchmark Results}

\paragraph{SportsMOT.} Table \ref{tab:sportsmot-results} presents the test set results, comparing {\proposed} with recent methods. Our method achieves the highest performance among all learning-based approaches, surpassing DiffMOT \cite{diffmot} by 1.06 HOTA, 1.6 IDF1, and 1.7 AssA. While our pipeline achieves the best HOTA score across both filter-based and learning-based models, it falls short of \cite{deep-eiou} in IDF1 and AssA.

\begin{table}[t]
\centering
\caption{\textbf{Quantitative tracking results on the SportsMOT test set.} The results marked with * use training and validation data for training the object detector. Best results are in \colorbox{rank_red}{Red}, second best in \colorbox{rank_orange}{Orange}.} 
\renewcommand{\arraystretch}{1.2} 
\vspace{-10px}
\resizebox{\linewidth}{!}{ 
\begin{tabular}{ll|ccccc}
    \hline
    & \textbf{Method} & \textbf{HOTA$\uparrow$} & \textbf{IDF1$\uparrow$} & \textbf{AssA$\uparrow$} & \textbf{MOTA$\uparrow$} & \textbf{DetA$\uparrow$} \\ 
    \midrule
    \multirow{8}{*}{\textit{\rotatebox{90}{Filter-based}}} 
    & FairMOT \cite{fairmot} & 49.3 & 53.5 & 34.7 & 86.4 & 70.2 \\ 
    & CenterTrack \cite{center-track} & 62.7 & 60.0 & 48.0 & 90.8 & 82.1 \\ 
    & ByteTrack \cite{bytetrack} & 62.8 & 69.8 & 51.2 & 94.1 & 77.1 \\ 
    & *ByteTrack \cite{bytetrack} & 64.1 & 71.4 & 52.3 & 95.9 & 78.5 \\ 
    & BoT-SORT \cite{botsort} & 68.7 & 70.0 & 55.9 & 94.5 & 84.4 \\ 
    & OC-SORT \cite{ocsort} & 71.9 & 72.2 & 59.8 & 94.5 & 86.4 \\
    & *OC-SORT \cite{ocsort} & 73.7 & 74.0 & 61.5 & 96.5 & 88.5 \\
    & Deep-EIoU \cite{deep-eiou} & \sbest{77.2} & \best{79.8} & \best{67.7} & 96.3 & 88.2 \\ 
    \midrule
    \multirow{11}{*}{\textit{\rotatebox{90}{Learning-based}}} 
    & QDTrack \cite{qdtrack} & 60.4 & 62.3 & 47.2 & 90.1 & 77.5 \\ 
    & GTR \cite{GTR} & 54.5 & 55.8 & 45.9 & 67.9 & 64.8 \\ 
    & TransTrack \cite{transtrack} & 68.9 & 71.5 & 57.5 & 92.6 & 82.7 \\ 
    & *MixSort-Byte \cite{sportsmot} & 65.7 & 74.1 & 54.8 & 96.2 & 78.8 \\ 
    & *MixSort-OC \cite{sportsmot} & 74.1 & 74.4 & 62.0 & 96.5 & 88.5 \\ 
    & MotionTrack \cite{motiontrack} & 74.0 & 74.0 & 61.7 & 96.6 & 88.8 \\ 
    & DiffMOT \cite{diffmot} & 72.1 & 72.8 & 60.5 & 94.5 & 86.0 \\ 
    & MambaTrack+ \cite{mambatrackexplore} & 71.3 & 71.1 & 58.6 & 94.9 & 86.7 \\  
    & MambaTrack \cite{mambatrack} & 72.6 & 72.8 & 60.3 & 95.3 & 87.6 \\  
    & *DiffMOT \cite{diffmot} & 76.2 & 76.1 & 65.1 & \best{97.1} & \sbest{89.3} \\ 
    & *ByteSSM \cite{trackssm} & 74.4 & 74.5 & 62.4 & 96.8 & 88.8 \\ 
    & \textbf{{\proposed}} (Ours) & \best{77.3} & \sbest{77.7} & \sbest{66.8} & \sbest{96.9} & \best{89.5} \\ 
    \hline
\end{tabular}
}
\label{tab:sportsmot-results}
\end{table}

\paragraph{VIP-HTD.} We benchmark the SOTA tracking methods on the VIP-HTD dataset to evaluate the performance of our method. Table \ref{tab:vip-htd-results} compares recent tracking methods with our approach, where {\proposed} achieves the highest HOTA score, surpassing both learning-based and filter-based models; demonstrating superior tracking performance.

\begin{table}[t]
\centering
\caption{\textbf{Quantitative tracking results for VIP-HTD ice hockey test set.} $\uparrow$ indicates that higher values are better. } 
\vspace{-10px}
\renewcommand{\arraystretch}{1.2}
\resizebox{\linewidth}{!}{ 
\begin{tabular}{ll|ccccc}
        \hline
        & \textbf{Method} & \textbf{HOTA}$\uparrow$ & \textbf{IDF1}$\uparrow$ & \textbf{AssA}$\uparrow$ & \textbf{MOTA}$\uparrow$ & \textbf{DetA}$\uparrow$ \\ 
        \midrule
        \multirow{3}{*}{\textit{\rotatebox{90}{\shortstack{Filter-\\based}}}}
        & ByteTrack \cite{bytetrack} & \sbest{64.4} & \best{81.1} & \best{64.8} & 73.9 & 64.2 \\ 
        &OC-SORT \cite{ocsort} & 61.0 & 75.4 & 58.9 & 74.6 & 
        63.4 \\
        &Deep OC-SORT \cite{ocsort} & 59.4 & 73.4 & 56.1 & 74.5 & 56.1 \\
        \midrule
        \multirow{3}{*}{\textit{\rotatebox{90}{\shortstack{Learning-\\based}}}} 
        &DiffMOT \cite{diffmot} & 64.1 & 79.4 & 63.6 & \sbest{76.1} & 65.0 \\ 
        &ByteSSM \cite{trackssm} & 63.4 & 77.7 & 61.8 & \best{76.2} & \sbest{65.4} \\ 
        &\textbf{{\proposed}} (Ours) & \best{65.1} & \sbest{80.1} & \sbest{64.6} & \best{76.2} & \best{65.9} \\ \hline
\end{tabular}
}
\label{tab:vip-htd-results}
\end{table}



\subsection{Ablation Study}

This section examines how key factors influence model performance, including the number of Mamba-Attention Blocks ($M$), tracklet window $(w)$, height adaptation in the association cost matrix, and buffer sizes ($b_1, b_2$) for EIoU. To ensure a focused analysis, training hyperparameters such as epochs, learning rate, and optimizers remain fixed.

\noindent \textbf{Tracklet length and Mamba-Attention blocks.} Table \ref{tab:ablation-window-block} presents an analysis of varying number of Mamba-Attention blocks $M$ and tracklet sequence length $w$. We report the HOTA scores in the SportsMOT dataset for this experiment. The results show the model maintains strong performance with $M = 4$ and $M = 5$ on SportsMOT with an 82.934 HOTA Score. Additionally, increasing $w$ enhances association, likely contributing to improved predictions with longer tracklet lengths. The best result is achieved with $(M = 4, w = 10)$. 

\noindent \textbf{Effect of varying buffer sizes $b_1$ and $b_2$.} Table \ref{tab:buffer-sizes-ablation} presents the impact of different buffer sizes $b_1$ and $b_2$ on tracker performance in SportsMOT, with height adaptation enabled throughout the analysis. The results indicate that the configuration $b_1 = 0.4$, $b_2 = 0.3$ achieves the highest HOTA (84.239) on the validation set, suggesting an optimal balance for accurate association.

\noindent \textbf{Impact of modified spatial association metric} Table \ref{tab:vip-htd-results} compares the effectiveness of EIoU and HA-IoU and HA-EIoU metrics against the standard IoU on the SportsMOT validation set. The results indicate that incorporating extended buffers (EIoU) enhances association performance. Furthermore, height adaptation combined with extended buffers (HA-EIoU) leads to an additional improvement, yielding $\approx$ 0.9 increase in HOTA and a $\approx$ 1.1 increase in IDF1 compared to EIoU, demonstrating its effectiveness in refining identity preservation and tracking accuracy in fast-motion scenarios such as team sports.

\begin{table}[t]
    \centering
    \caption{\textbf{HOTA Scores for different combinations of block sizes ($M$) and window size ($w$) on SportsMOT validation set.} We use the standard IoU without buffer sizes $b_1$ and $b_2$ for both spatial association update stages. }
    \vspace{-10px}
    \renewcommand{\arraystretch}{1.0}

    \small 
    \begin{tabular}{l|ccccc}
        \toprule
         & \textbf{M = 2} & \textbf{M = 3} & \textbf{M = 4} & \textbf{M = 5} \\
        \midrule
        \textbf{w = 5} & 82.928 & 82.384 & 82.561 & 82.407 \\
        \textbf{w = 7} & 82.555 & 82.447 & 82.626 & 82.858 \\
        \textbf{w = 10} & 82.581 & 82.488 & \best{82.934} & 82.696 \\
        \textbf{w = 12} & 82.661 & 82.461 & 82.243 & 82.773 \\
        \textbf{w = 15} & 82.695 & 82.658 & 82.881 & 82.881 \\
        \bottomrule
    \end{tabular}
    \label{tab:ablation-window-block}
\end{table}

\begin{table}[t]
    \centering
    \caption{\textbf{Effect of different buffer sizes $b_1$ and $b_2$ on SportsMOT validation set.}} 
    \vspace{-10px}
    \label{tab:buffer-sizes-ablation}
    \renewcommand{\arraystretch}{1.0}
    \small
\resizebox{\linewidth}{!}{ 
    \begin{tabular}{c|ccccc}
        \toprule
        $b_1 \setminus b_2$ & \textbf{0.25} & \textbf{0.30} & \textbf{0.35} & \textbf{0.40} & \textbf{0.45} \\
        \midrule
        \textbf{0.25} & 83.869 & 83.885 & 83.769 & 83.769 & 83.775 \\
        \textbf{0.30} & 83.937 & 83.892 & 83.787 & 83.780 & 83.784 \\
        \textbf{0.35} & 83.937 & 83.937 & 83.839 & 83.837 & 83.836 \\
        \textbf{0.40} & 83.974 & \best{84.239} & 83.865 & 83.866 & 83.868 \\
        \hline
    \end{tabular}}
\end{table}

\begin{table}[t]
\centering
\caption{\textbf{Performance comparison of effect of different spatial association metric for SportsMOT validation set.}} 
\vspace{-10px}
\renewcommand{\arraystretch}{1.0}
\resizebox{\linewidth}{!}{ 
\begin{tabular}{l|ccccc}
        \hline
         & \textbf{HOTA}$\uparrow$ & \textbf{DetA}$\uparrow$ & \textbf{AssA}$\uparrow$ & \textbf{MOTA}$\uparrow$ & \textbf{IDF1}$\uparrow$ \\ 
         \midrule
        IoU & 82.934	& 94.073	& 73.134	& 98.499	
        & 82.429 \\
        EIoU & 83.316	& 94.108	& 73.783	& \best{98.693}	& 82.862 \\       
        HIoU & 83.421 & 	94.113	&  73.964  & 	98.466	& 83.024 \\ \hline
        HA-EIoU & \best{84.239} & 	\best{94.165}	&  \best{75.379}  & 	98.692	& \best{83.931} \\\hline
\end{tabular}
}
\label{tab:vip-htd-results}
\end{table}

\begin{figure*}
    \centering
        \includegraphics[width = \linewidth]{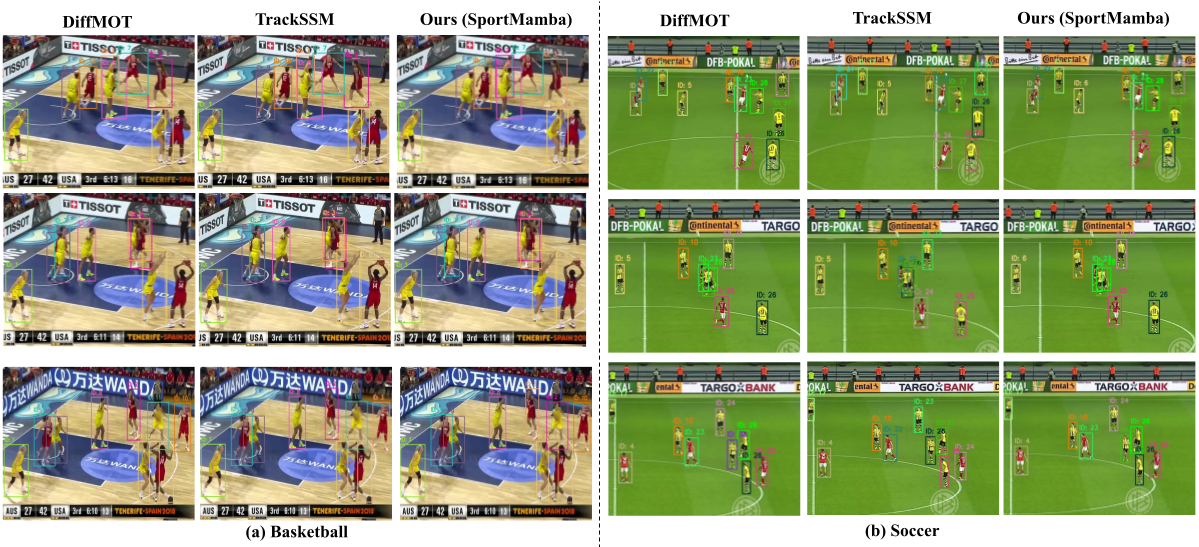}  
    \vspace{-20px}
    \caption{\textbf{Qualitative comparison of {\proposed} in SportsMOT dataset on two categories: a) basketball and b) soccer.}}
    \label{fig:sportmamba-qual}
    \vspace{-10px}
\end{figure*}

\begin{figure}[t]
    \centering
        \includegraphics[width = 0.49\textwidth]{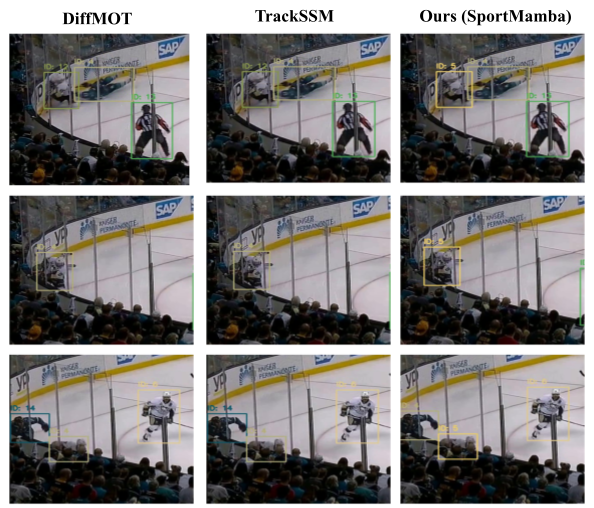}  
    \vspace{-20px}
    \caption{\textbf{Qualitative comparison of {\proposed} in VIT-HTD dataset. }}
    \label{fig:comparison-viphtd}
    \vspace{-10px}
\end{figure}

\subsection{Qualitative Analysis}
Figures \ref{fig:sportmamba-qual} and \ref{fig:comparison-viphtd} provide a qualitative comparison of {\proposed} against recent data-driven motion models, DiffMOT \cite{diffmot} and TrackSSM \cite{trackssm}. Each figure consists of three rows: 1) The first row presents a frame before any occlusion occurs; 2) Second row highlights a frame where occlusion or missed detections take place in different sequences; and 3) Third row demonstrates how each tracker re-associates player identities after occlusion.

In Figure \ref{fig:comparison-viphtd}, a player becomes occluded in the second frame and is incorrectly assigned a new ID in DiffMOT and TrackSSM when reappearing, whereas {\proposed} successfully retains the correct ID. Similarly, in Figure \ref{fig:sportmamba-qual} (left), players ID 3, ID 2, and ID 10 overlap. DiffMOT and TrackSSM mistakenly change ID 3 to ID 11, while {\proposed} accurately handles the occlusion and preserves identity consistency. Similarly, in Soccer (Figure \ref{fig:sportmamba-qual} (right), both TrackSSM and {\proposed} successfully reassign player identities after occlusion, unlike DiffMOT. These results highlight {\proposed}’s superior ability to maintain identity consistency despite occlusions.
\section{Conclusion}

In this work, we introduced {\proposed}, an adaptive MOT model designed to tackle the inherent challenges of tracking fast-moving players in team sports. To maintain computational efficiency, {\proposed} integrates SSM with the self-attention mechanism to capture the non-linear motion patterns from tracklet sequences. Upon extracting the predicted motion, we introduced a hybrid association strategy that combines appearance-based features with an introduced height-adaptive spatial association metric with extended buffers (HA-EIoU). Experimentation on the SportMOT dataset comprising of three sports categories (basketball, soccer, volleyball) demonstrated SOTA performance of {\proposed} in key metrics, including HOTA, IDF1, AssA, and DetA. To assess the generalizability of {\proposed}, we evaluated it on the VIP-HTD dataset in a zero-shot setting, where it achieved superior performance across all metrics, showcasing its effectiveness in fast-paced team sports. This highlights {\proposed}'s generalizability to diverse sports settings. While {\proposed} excels in motion prediction and data association, we acknowledge that severe motion blur can lead to missed detections and weakened appearance cues, resulting in broken tracklets. Nevertheless, {\proposed} provides a strong foundation for advancing MOT for team sports, given its ability to tackle the prevalent challenges of dynamic team sports.

\vspace{-5px}
\paragraph{Acknowledgement}
This work was supported in part by Stathletes, the Natural Sciences and Engineering
Research Council of Canada and MITACS.
{
    \small
    \bibliographystyle{ieeenat_fullname}
    \bibliography{main}
}


\end{document}